\documentclass[10pt,twocolumn]{article}

\usepackage{cvpr}
\usepackage{times}
\usepackage{graphicx}
\usepackage{amsmath}
\usepackage{amssymb}
\usepackage{setspace}

\usepackage[pagebackref=true,breaklinks=true,colorlinks,bookmarks=false]{hyperref}

\usepackage{framed,multirow}

\usepackage{latexsym}
\usepackage{pifont}
\usepackage{caption}
\usepackage[page]{appendix}

\newcommand{\diag}{\mathop{\mathrm{diag}}\nolimits}
\newcommand{\xmark}{\text{\ding{55}}}

\cvprfinalcopy 

\ifcvprfinal\fi
\begin{document}

\title{A neural network based on SPD manifold learning\\for skeleton-based hand gesture recognition}
\author{Xuan Son Nguyen, Luc Brun, Olivier L\'ezoray, S\'ebastien Bougleux\\
Normandie Univ, UNICAEN, ENSICAEN, CNRS, GREYC, 14000 Caen, France\vspace{0.2em}\\
{\small Presented at the IEEE International Conference on Computer Vision and Pattern Recognition (CVPR)}\vspace{-0.2em}\\
{\small 2019, June 16-21, Long Beach, CA (\url{http://cvpr2019.thecvf.com/})}
\vspace{-1.2em}
}

\maketitle

\begin{abstract}
This paper proposes a new neural network based on SPD manifold learning for skeleton-based hand gesture recognition. Given the stream of hand's joint positions, our approach combines two aggregation processes on respectively   spatial and  temporal domains. The pipeline of our  network architecture consists in three main stages. 
The first stage is based on a convolutional layer to increase the discriminative power of learned features. The second stage relies on different architectures for spatial and temporal Gaussian aggregation of joint features. The third stage learns a final SPD matrix from skeletal data. A new type of layer is proposed for the third stage, based on a variant of stochastic gradient descent on Stiefel manifolds. The proposed network is validated on two challenging datasets and shows state-of-the-art accuracies on both datasets.
\end{abstract}
\vspace{-0.2em}
\section{Introduction}
\label{sec:intro}

Hand gesture recognition is an important research topic with  applications in many fields, e.g., assisted living, human-robot interaction or sign language interpretation.
A large family of hand gesture recognition methods is based on low-level features extracted from images, e.g., spatio-temporal interest points. 
However, with the introduction of affordable depth sensing cameras, e.g., Intel Realsense or Microsoft Kinect, and the availability of
highly accurate joint tracking algorithms, skeletal data can be obtained  effectively with good precision. Skeletal data provides a rich and high level description of the hand. This has led to an extensive development of approaches for skeleton-based hand gesture recognition in recent years. 

Early works on the recognition of hand gestures or human actions from skeletal data are based on a modeling of the skeleton's movement as  time series~\cite{LieGroup14,ActionletEns12}. The recognition step is thus based on the comparison of sequences of features describing skeleton's movements using, e.g.,  Dynamic Time Warping~\cite{LieGroup14} or Fourier Temporal Pyramid~\cite{ActionletEns12}. 

Such approaches ignore the high correlations existing between the movement of two adjacent hand joints (e.g., two joints of a same finger) within a hand gesture. Taking into account this information is a crucial step for hand gesture recognition and  requires the definition and the appropriate processing of hand joints' neighborhoods. 

Recently, graph convolutional networks for action recognition~\cite{LiGraphConvAAAI18,YanAAAI18} have shown excellent performance 
by taking into account physical connections of body joints defined by the underlying structure of body skeleton. 
While the use of physical connections of skeleton joints are important for capturing discriminating cues in hand gesture and action recognition,
the identification of other connections induced by the performed gestures and actions are also useful and can greatly improve recognition accuracy~\cite{ShiArXiv18}.

Motivated by this observation, we model in this work the hand
skeleton as a 2D grid where connections, different from the classical physical connections of hand joints, are added to
better capture patterns defined by  hand joints' movements. 
Figure~\ref{fig:hand_graph}(a) shows the hand joint positions estimated by an Intel Realsense camera. 
Since the hand skeleton has an irregular geometric structure that differs from grid-shaped structures,
the 2D grid is constructed from the hand skeleton by removing some hand joints and adding connections between neighboring joints.  
Figure~\ref{fig:hand_graph}(b) shows a 2D grid corresponding to the hand skeleton in Fig.~\ref{fig:hand_graph}(a). 
This 2D grid integrates adjacency relationships between hand joints that often have correlated movements. Moreover, this modeling allows us to use a classical convolutional layer instead of a graph convolutional operator on an arbitrary geometric graph~\cite{YanAAAI18}.

Our approach relies on SPD matrices to aggregate features resulting from the convolutional
layer. The SPD matrices considered in this work combine  mean and covariance information which have been shown effective in various vision tasks~\cite{Harandi2018DimensionalityRO,HuangGeoAware18,HuangLogMeT2015}. 
Since SPD matrices are known to lie on a Riemannian manifold, specific layers for deep neural networks of SPD matrices should be designed~\cite{HuangGool17,ZhangSPDNet17}.
Beside  good performances on action recognition tasks, these networks 
do not put a focus on  spatial and temporal relationships of skeleton joints. This motivates us to design a neural network model 
for learning a SPD matrix-based gesture representation from skeletal data with a special attention on  those  relationships.  
In our work, the encoding of spatial and temporal relationships of hand joints is performed using different network architectures. This allows to capture relevant statistics for individual hand joints as well as groups of hand joints
whose movement is highly correlated with that of other joints in the group. The experimental evaluation shows that our method significantly improves the state-of-the-art methods on two standard datasets.

\section{Related Works}\label{sec:related_work}
This section presents representative works for skeleton-based hand gesture recognition (Sec.~\ref{subsec:ske_approaches}) and deep neural networks for SPD manifold learning (Sec.~\ref{subsec:deep_learning_spd}).
\subsection{Skeleton-Based Gesture Recognition}
\label{subsec:ske_approaches}

Most of approaches can be categorized as hand-crafted feature-based approaches or deep learning approaches. Hand-crafted feature-based approaches describe relationships of hand and body joints in different forms to represent gestures and actions. The  simplest proposed relationships are relative positions between pairs of joints~\cite{Luo13,SmedtCVPRW16,EigenJoints}. More complex relationships were also exploited, e.g., skeletal quad~\cite{SkeQuad14} or 3D geometric relationships of body parts in a Lie group~\cite{LieGroup14}. Temporal relationships have also been taken into account and proven effective~\cite{PoseSet13}. While all joints are involved in the performed gestures and actions, only a subset of key joints is important for the recognition task.
These are called informative joints and they can be automatically  identified using information theory~\cite{OFLI201424}. 
This allows to avoid considering non-informative joints that often bring noise 
and degrade performance.
 
Motivated by the success of deep neural networks in various vision tasks~\cite{Girshick15RCNN,HeResNet16,Krizhevsky12ImageNet},
deep learning approaches for action and gesture recognition have been extensively studied in recent years. 
To capture spatial and temporal relationships of hand and body joints, they rely mainly on 
Convolutional Neural Network (CNN)~\cite{DevineauFG18,KeNewReRecCVPR17,LIUEnhViewPR2017,LiuEvoMapsCVPR18,Nez2018,Wang16TrajMapsCNN},
Recurrent Neural Network (RNN)~\cite{NN15,Wang2017ModelingTD} and Long Short-Term Memory (LSTM)~\cite{LiuCVPR17LSTM,Nez2018,Shahroudy16NTU}. While hand-crafted feature-based approaches have used informative joints to improve recognition accuracy, deep learning approaches were based on attention mechanism to selectively focus on relevant parts of skeletal data~\cite{LiuTrustGateECCV16,WengTraversalConvECCV18}. 
Recently, deep learning on manifolds and graphs has increasingly attracted attention. Approaches following this line of research have also been successfully applied to skeleton-based 
action recognition~\cite{HuangGool17,Huang17DLLieGroup,HuangAAAI18,LiGraphConvAAAI18,YanAAAI18}. 
By extending classical operations like convolutions to manifolds and graphs while respecting the underlying geometric structure of data, they have demonstrated superior performance over other approaches.

\subsection{Deep Learning of SPD Matrices}
\label{subsec:deep_learning_spd}

In recent years the deep learning community has shifted its focus towards developing approaches that deal with data in a non-Euclidean domain, e.g., Lie groups~\cite{Huang17DLLieGroup}, SPD manifolds~\cite{HuangGool17} or Grassmann manifolds~\cite{HuangAAAI18}. Among them, those that deal with SPD manifolds have received particular attention. This comes from the popular applications of SPD matrices in many vision problems~\cite{Bilinski15,Guo13,Harandi2013,Yuan09}.

Deep neural networks for SPD matrix learning aim at projecting a high-dimensional SPD matrix into a more discriminative low-dimensional one. Differently from classical CNNs, their layers are designed so that they preserve the geometric structure of input SPD matrices, i.e., their output are also SPD matrices. 
In~\cite{DongSPD17}, a 2D fully connected layer was proposed for the projection, while in~\cite{HuangGool17} it was achieved by a Bimap layer. Inspired by ReLU layers in CNNs, different types of layers that perform non-linear transformations of SPD matrices were also introduced~\cite{DongSPD17,EnginKspd17,HuangGool17}. 
To classify the final SPD matrix, a layer is generally required to map it to an Euclidean space. Most of approaches rely on the two widely used operations in many machine learning models, i.e., singular value decomposition (SVD) and eigen value decomposition (EIG) for constructing this type of layers~\cite{HuangGool17,LiIsSecond17,G2DeNet17}. As  gradients involved in SVD and EIG cannot be computed by traditional backpropagation, they exploit the chain rule established by Ionescu et al.~\cite{Ionescu2015} for backpropagation of matrix functions in deep learning.

\begin{figure}[t]
{
\def\a{$w_1$}
\def\b{$w_2$}
\def\c{$w_4$}
\def\d{$w_6$}
\def\e{$w_8$}
\def\f{$w_3$}
\def\g{$w_5$}
\def\h{$w_7$}
\def\i{$w_9$}
\centerline{\scalebox{0.45}{\input{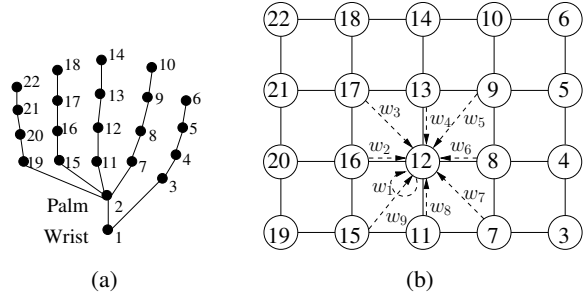}}}
\caption{\label{fig:hand_graph} (a) Hand joints estimated by an Intel RealSense camera  
(b) graph of hand skeleton and the weights associated with the neighbors of node 12 in the convolutional layer.} 
}
\end{figure}

\section{The Proposed Approach}
\label{sec:proposed_method_deep}

In this section, we present our  network model referred to as Spatial-Temporal and Temporal-Spatial Hand Gesture Recognition Network (ST-TS-HGR-NET). 
An overview of our network is given in Section~\ref{subsec:overview}. The  different components of our network
are explained in Sections~\ref{subsec:convolutional_layer},~\ref{subsec:spatial_temporal_network},~\ref{subsec:temporal_spatial_network}, 
and~\ref{subsec:classification_network}. In Section~\ref{subsec:gesture_recognition}, we show how our network is trained for gesture recognition.
Finally, Section~\ref{subsec:links_previous_works} points out the relations of our approach with previous approaches. 

\subsection{Overview of The Proposed Network}
\label{subsec:overview}

\begin{figure}[t]
{
\centerline{\scalebox{0.23}{\input{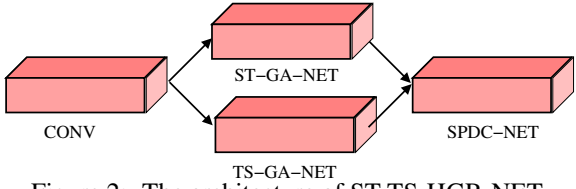}}}
\caption{\label{fig:combined_network} The architecture of ST-TS-HGR-NET. 
} 
}
\end{figure}

Our network illustrated in Fig.~\ref{fig:combined_network} is made up of three components. 
The first component, referred to as CONV, is a convolutional layer applied on the 2D grid encoding the  hand skeletal data (Fig.~\ref{fig:hand_graph}).  Filter weights are shared over all frames of the sequence. 

The second component is based on  the Gaussian embedding method of~\cite{Lovric00} and is used to capture  first- and second-order statistics. This component is composed of  two different architectures for feature aggregation referred to as Spatial-Temporal Gaussian Aggregation Sub-Network (ST-GA-NET) and Temporal-Spatial Gaussian Aggregation Sub-Network (TS-GA-NET).

The third component, referred to as SPD Matrix Learning and Classification Sub-Network (SPDC-NET), 
learns a SPD matrix from a set of SPD matrices and maps the resulting SPD matrix,
which lies on a Riemannian manifold, to an Euclidean space for classification. 

In the following, we explain in detail each component of our network. 
The backpropagation procedures of our network's layers are given in Appendix~\ref{appendix}.

\subsection{Convolutional Layer}
\label{subsec:convolutional_layer}

\begin{figure}[t]
  \begin{center}
    \begin{tabular}{c}
      \includegraphics[width=1.0\linewidth, trim = 200 330 170 310, clip=true]{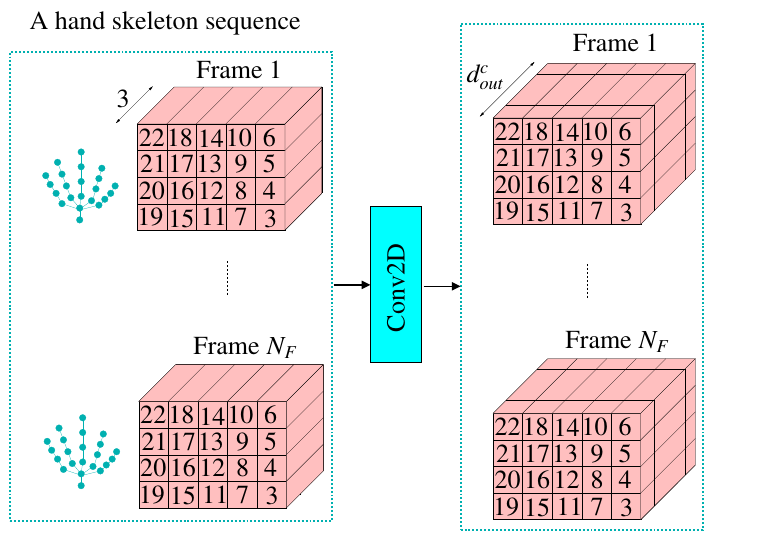} 
    \end{tabular}
  \end{center} 
\caption{\label{fig:conv_network} Illustration of sub-network CONV. }
\end{figure}

The convolutional layer (Fig.~\ref{fig:conv_network})  used in the first place of our network allows to combine joints with correlated variations (Section~\ref{sec:intro}). Let $N_J$ and $N_F$ be respectively the number of hand joints and the length of the skeleton sequence. Let us denote by $\mathbf{p}^t_{0,i} \in \mathbb{R}^3,i=1,\ldots,N_J,t=1,\ldots,N_F$, the 3D coordinates of hand joint $i$ at frame $t$. 
We define a 2D grid where each node represents a hand joint $i$ at a frame $t$ (Section~\ref{sec:intro}). 
The grid has three channels corresponding to the x, y, and z coordinates of hand joints. 
Fig.~\ref{fig:hand_graph}(b) shows the 2D grid corresponding to the hand skeleton in Fig.~\ref{fig:hand_graph}(a), 
where each node has at most 9 neighbors including itself. Let $d_{out}^{c}$ be the output dimension  of the convolutional layer. 
Let us denote by $\mathbf{p}^t_{i} \in \mathbb{R}^{d_{out}^{c}},i=3,\ldots,N_J,t=1,\ldots,N_F$, the output of the convolutional layer. 
The output feature vector at node $i$ is computed as:
\begin{equation}\label{eq:graph_convolution}
\mathbf{p}_{i}^t = \sum_{j \in \mathcal{N}_i} \mathbf{W}_{l(j,i)} \mathbf{p}_{0,j}^t,
\end{equation}
where $\mathcal{N}_i$ is the set of neighbors of node $i$, $\mathbf{W}_{l(j,i)}$ is the filter weight matrix, 
and $l(j,i)$ is defined as:
\begin{equation}\label{eq:vertex_weight_label}
{\small{\begin{array}{|c|c|c|c|c|c|c|c|c|c|}
    \hline
     j-i    & 0 & 4 & 5 & 1 & -3 & -4 & -5 & -1 & 3 \\\hline
     l(j,i) & 1 & 2 & 3 & 4 & 5 & 6 & 7 & 8 & 9\\\hline
\end{array}}}
\end{equation}

\subsection{Spatial-Temporal Gaussian Aggregation Sub-Network}
\label{subsec:spatial_temporal_network}

\begin{figure}[t]
  \begin{center}
    \begin{tabular}{c}
      \includegraphics[width=1.2\linewidth, trim = 50 280 260 250, clip=true]{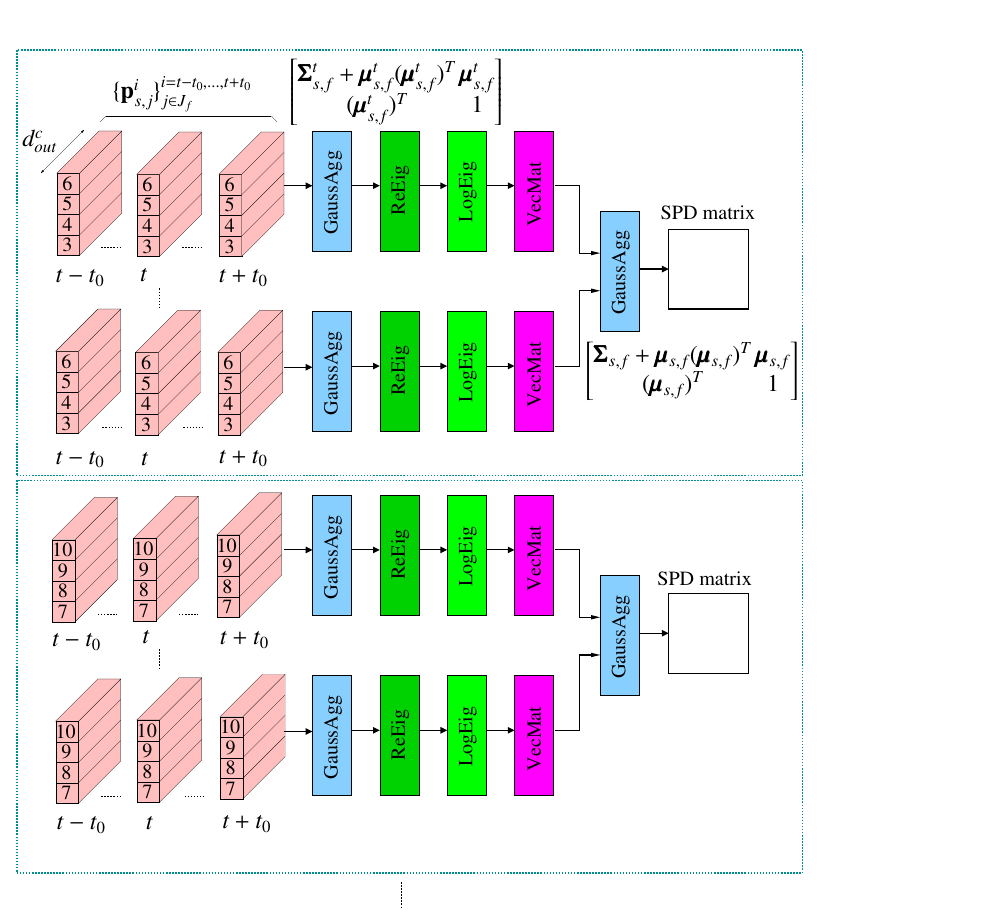}
    \end{tabular}
  \end{center} 
\caption{\label{fig:st_ga_network} The architecture of ST-GA-NET (two different branches are shown).}
\end{figure}

To capture the temporal ordering of a skeleton sequence, a number of sub-sequences are constructed  and then fed to different branches
of ST-GA-NET (see Fig.~\ref{fig:st_ga_network}). A branch of ST-GA-NET is designed to aggregate features for a sub-sequence of a specific finger. 
In this paper, we construct six sub-sequences for each skeleton sequence. 
The first sub-sequence is the original sequence. The next two sub-sequences are obtained by dividing the sequence into two sub-sequences of equal length.
The last three sub-sequences are obtained by dividing the sequence into three sub-sequences of equal length.
This results in $30$ branches for ST-GA-NET ($6$ sub-sequences $\times$ $5$ fingers).


To aggregate features in a branch associated with sub-sequence $s$ and finger $f$, $s=1,\ldots,6,f=1,\ldots,5$, each frame of sub-sequence $s$
is processed through $4$ layers.  
Let $J_f$ be the set of hand joints belonging to finger $f$, 
$t^s_b,t^s_e$ be the beginning and ending frames of sub-sequence $s$, $t$ be a given frame of sub-sequence $s$, $\{ \mathbf{p}_{s,j}^i | j \in J_f,i=t^s_b,\ldots,t^s_e \}$
be the subset of output feature vectors of the convolutional layer that are fed to the branch. 
Let us finally consider a sliding window $\{t-t_0,\dots,t+t_0\}$  centered on frame $t$. Following previous works~\cite{Li17,SERRA201522}, we assume that $\mathbf{p}_{s,j}^i$, $j \in J_f,i=t-t_0,\ldots,t+t_0$, are independent 
and identically distributed samples from a Gaussian distribution (hereafter abbreviated as Gaussian for simplicity):

\begin{align}
\begin{split}
\mathcal{N}(& \mathbf{p};\pmb{\mu}^t_{s,f},\pmb{\Sigma}^t_{s,f}) = \\ & \frac{1}{|2\pi \pmb{\Sigma}^t_{s,f}|^{\frac{1}{2}}} \exp (-\frac{1}{2}(\mathbf{p}-\pmb{\mu}^t_{s,f})^T (\pmb{\Sigma}^t_{s,f})^{-1} (\mathbf{p}-\pmb{\mu}^t_{s,f})),
\end{split}
\end{align}
where $|.|$ is the determinant, $\pmb{\mu}^t_{s,f}$ is the mean vector and $\pmb{\Sigma}^t_{s,f}$ is the covariance matrix. 
The parameters of the Gaussian can be estimated as:

\begin{equation}
\pmb{\mu}^t_{s,f} = \frac{1}{(2t_0+1)|J_f|} \sum_{j \in J_f} \sum_{i=t-t_0}^{t+t_0} \mathbf{p}_{s,j}^i,
\end{equation}

\begin{equation}
\pmb{\Sigma}^t_{s,f} = \frac{1}{(2t_0+1)|J_f|} \sum_{j \in J_f} \sum_{i=t-t_0}^{t+t_0} (\mathbf{p}_{s,j}^i - \pmb{\mu}^t_{s,f})(\mathbf{p}_{s,j}^i - \pmb{\mu}^t_{s,f})^T.
\end{equation}
Based on the method in~\cite{Lovric00} that embeds the space of Gaussians in the Riemannian symmetric space, 
the Gaussian can be identified as a SPD matrix given by:  

\begin{equation}\label{eq:spd_elementary}
\mathbf{Y}^t_{s,f} = \begin{bmatrix}  \pmb{\Sigma}^t_{s,f} + \pmb{\mu}^t_{s,f}(\pmb{\mu}^t_{s,f})^T & \pmb{\mu}^t_{s,f} \\ (\pmb{\mu}^t_{s,f})^T & 1 \end{bmatrix}.
\end{equation}

The GaussAgg layer is designed to perform the computation of Eq.~\ref{eq:spd_elementary}, that is:

\begin{equation}\label{eq:gaussagg_mapping}
\mathbf{Y}^t_{s,f} = h_{ga}(\{\mathbf{p}_{s,j}^i\}_{j \in J_f}^{i=t-t_0,\ldots,t+t_0}),
\end{equation}
where $h_{ga}$ is the mapping of the GaussAgg layer, $\mathbf{Y}^t_{s,f}$ is the output of the GaussAgg layer. 

\begin{figure*}[t]
  \begin{center}
    \begin{tabular}{c}
      \includegraphics[width=1.0\linewidth, trim = 50 345 30 310, clip=true]{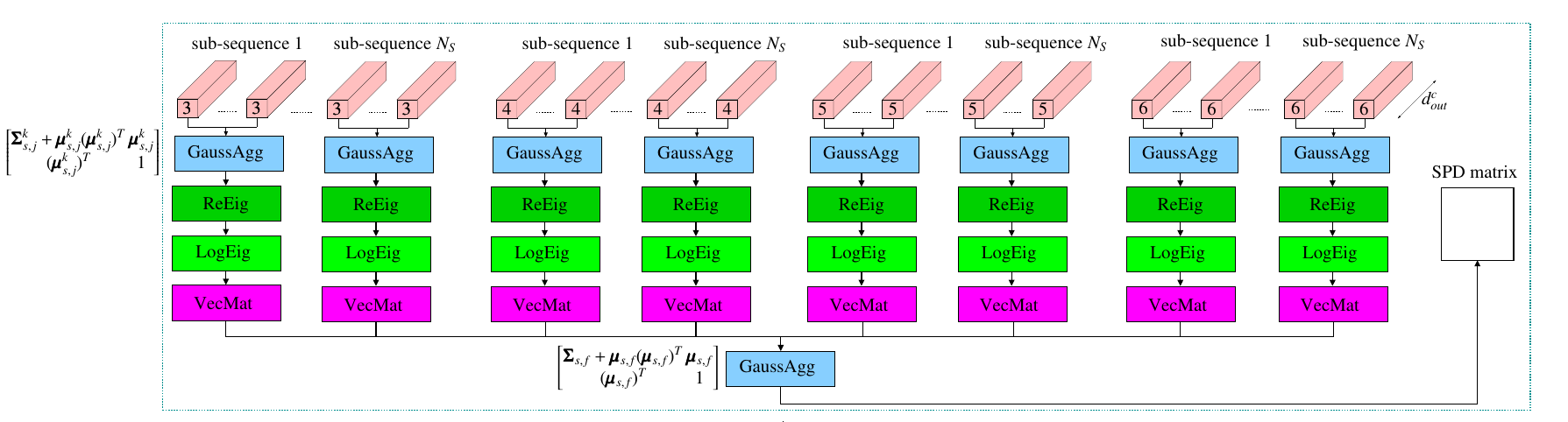} 
    \end{tabular}
  \end{center} 
\caption{\label{fig:ts_ga_network} Illustration of a branch of TS-GA-NET.}
\end{figure*}

The next layer ReEig~\cite{HuangGool17} introduces non-linear transformations of SPD matrices via a mapping defined as:

\begin{equation}\label{eq:reeig}
\mathbf{Y}^t_{s,f} = h_r (\mathbf{X}^t_{s,f}) = \mathbf{U} \max(\epsilon \mathbf{I}, \mathbf{V}) \mathbf{U}^T,
\end{equation}
where $h_r$ is the mapping of the ReEig layer, $\mathbf{X}^t_{s,f}$ and $\mathbf{Y}^t_{s,f}$ are the input and output SPD matrices, 
$\mathbf{X}^t_{s,f} = \mathbf{U} \mathbf{V} \mathbf{U}^T$
is the eigen-decomposition of $\mathbf{X}^t_{s,f}$, 
$\epsilon$ is a rectification threshold, $\mathbf{I}$ is the identity matrix, $\max (\epsilon \mathbf{I}, \mathbf{V})$
is a diagonal matrix whose diagonal elements are defined as:

\begin{equation}\label{eq:a_definition}
(\max (\epsilon \mathbf{I}, \mathbf{V}))(i,i) = \left\lbrace\begin{array}{ll} \mathbf{V}(i,i) & \text{if } \mathbf{V}(i,i) > \epsilon \\ \epsilon & \text{if } \mathbf{V}(i,i) \le \epsilon. \end{array}\right.
\end{equation}

After the ReEig layer, the LogEig layer~\cite{HuangGool17} is used to map SPD matrices to Euclidean spaces. 
Formally, the mapping of this layer is defined as:
\begin{equation}
\mathbf{Y}^t_{s,f} = h_l (\mathbf{X}^t_{s,f}) = \log (\mathbf{X}^t_{s,f}) = \mathbf{U} \log(\mathbf{V}) \mathbf{U}^T,
\end{equation}
where $h_l$ is the mapping of the LogEig layer, $\mathbf{X}^t_{s,f}$ and $\mathbf{Y}^t_{s,f}$ are the input and output SPD matrices, as before.

The next layer, referred to as VecMat, vectorizes SPD matrices by the following mapping~\cite{Tuzel08}:

\begin{align}
\begin{split}
& \mathbf{y}^t_{s,f} = h_{vm} (\mathbf{X}^t_{s,f}) = [\mathbf{X}^t_{s,f}(1,1), \sqrt{2}\mathbf{X}^t_{s,f}(1,2), \ldots, \\ & \sqrt{2}\mathbf{X}^t_{s,f}(1,d^c_{out}+1), \mathbf{X}^t_{s,f}(2,2), \sqrt{2}\mathbf{X}^t_{s,f}(2,3),\ldots, \\ & \mathbf{X}^t_{s,f}(d^c_{out}+1,d^c_{out}+1)]^T,
\end{split}
\end{align}
where $h_{vm}$ is the mapping of the VecMat layer, $\mathbf{X}^t_{s,f} \in \mathbb{R}^{d^c_{out}+1}$ is the input matrix, $\mathbf{y}^t_{s,f}$ is the output vector, $\mathbf{X}^t_{s,f}(i,i),i=1,\ldots,d^c_{out}+1$, are the diagonal entries of $\mathbf{X}^t_{s,f}$ and $\mathbf{X}^t_{s,f}(i,j),i<j,i,j=1,\ldots,d^c_{out}+1$, are the off-diagonal entries.

We again assume that $\mathbf{y}^t_{s,f},t=t^s_b,\ldots,t^s_e$, are independent and identically distributed
samples from a Gaussian $\mathcal{N}(\mathbf{y};\pmb{\mu}_{s,f},\pmb{\Sigma}_{s,f})$ whose parameters can be estimated as:

\begin{align}
\pmb{\mu}_{s,f} &= \frac{1}{t^s_e-t^s_b+1} \sum_{t=t^s_b}^{t^s_e} \mathbf{y}^t_{s,f},\\
\pmb{\Sigma}_{s,f} &= \frac{1}{t^s_e-t^s_b+1} \sum_{t=t^s_b}^{t^s_e} (\mathbf{y}^t_{s,f} - \pmb{\mu}_{s,f})(\mathbf{y}^t_{s,f} - \pmb{\mu}_{s,f})^T.
\end{align}
The second GaussAgg layer then performs the mapping:

\begin{align}
\begin{split}
\mathbf{Y}_{s,f} & = h_{ga}(\{ \mathbf{y}^t_{s,f} \}^{t=t^s_b,\ldots,t^s_e}) \\ & =  \begin{bmatrix}  \pmb{\Sigma}_{s,f} + \pmb{\mu}_{s,f}(\pmb{\mu}_{s,f})^T & \pmb{\mu}_{s,f} \\ (\pmb{\mu}_{s,f})^T & 1 \end{bmatrix}.
\end{split}
\end{align}
The resulting SPD matrix $\mathbf{Y}_{s,f}$ describes variations of finger $f$ along sub-sequence $s$.

\subsection{Temporal-Spatial Gaussian Aggregation Sub-Network}
\label{subsec:temporal_spatial_network}
Similarly to ST-GA-NET, TS-GA-NET is composed of 30 branches where each branch aggregates features for a sub-sequence of a specific finger.
The sub-sequences are constructed in exactly the same way as ST-GA-NET. 
However, the feature aggregation procedure at the first and second GaussAgg layers are performed differently.
More precisely, considering the branch associated with sub-sequence $s$ and finger $f$. First, sub-sequence $s$ is further divided into  
$N_S$ sub-sequences of equal length. Let $t^s_{b,k}$ and $t^s_{e,k}$, $k=1,\ldots,N_S$, be the beginning and ending frames of these sub-sequences. 
Then for a given hand joint $j \in J_f$ and sub-sequence $k$, the first GaussAgg layer computes a SPD matrix given as:

\begin{equation}\label{eq:spd_elementary_ts}
\mathbf{Y}^k_{s,j} = \begin{bmatrix}  \pmb{\Sigma}^k_{s,j} + \pmb{\mu}^k_{s,j}(\pmb{\mu}^k_{s,j})^T & \pmb{\mu}^k_{s,j} \\ (\pmb{\mu}^k_{s,j})^T & 1 \end{bmatrix},
\end{equation}
where $\pmb{\mu}^k_{s,j} = \frac{1}{t^s_{e,k}-t^s_{b,k}+1} \sum_{t=t^s_{b,k}}^{t^s_{e,k}} \mathbf{p}_{s,j}^t$ and 
$\pmb{\Sigma}^k_{s,j} = \frac{1}{t^s_{e,k}-t^s_{b,k}+1} \sum_{t=t^s_{b,k}}^{t^s_{e,k}} (\mathbf{p}_{s,j}^t - \pmb{\mu}^k_{s,j})(\mathbf{p}_{s,j}^t - \pmb{\mu}^k_{s,j})^T$.

Note that $\mathbf{Y}^k_{s,j}$ encodes the first- and second-order statistics of hand joint $j$ computed within sub-sequence $k$. This temporal variation of individual joints is not captured by the first GaussAgg layer of ST-GA-NET. The resulting SPD matrices are processed through the ReEig, LogEig and VecMat layers.
Let $\mathbf{y}^k_{s,j},k=1,\ldots,N_S,j \in J_f$, be the output vectors of the VecMat layer of the branch. The second GaussAgg layer of TS-GA-NET then performs the following mapping:

\begin{align}
\begin{split}
\mathbf{Y}_{s,f} & = h_{ga}(\{ \mathbf{y}^k_{s,j} \}^{k=1,\ldots,N_S}_{j \in J_f}) \\ & =  \begin{bmatrix}  \pmb{\Sigma}_{s,f} + \pmb{\mu}_{s,f}(\pmb{\mu}_{s,f})^T & \pmb{\mu}_{s,f} \\ (\pmb{\mu}_{s,f})^T & 1 \end{bmatrix},
\end{split}
\end{align}
where $\pmb{\mu}_{s,f}$ and $\pmb{\Sigma}_{s,f}$ can be estimated as:

\begin{align}
&\pmb{\mu}_{s,f} = \frac{1}{N_S|J_f|} \sum_{j \in J_f} \sum_{k=1}^{N_S} \mathbf{y}^k_{s,j},\\
&\pmb{\Sigma}_{s,f} = \frac{1}{N_S|J_f|} \sum_{j \in J_f} \sum_{k=1}^{N_S} (\mathbf{y}^k_{s,j} - \pmb{\mu}_{s,f})(\mathbf{y}^k_{s,j} - \pmb{\mu}_{s,f})^T.
\end{align}

\subsection{SPD Matrix Learning and Classification Sub-Network} 
\label{subsec:classification_network}

The outputs of sub-networks ST-GA-NET and TS-GA-NET are sets of SPD matrices. The objective of the classification sub-network (see Fig.~\ref{fig:spdc_network})
is to transform those sets to a new SPD matrix, then map it to an Euclidean space for classification. 
The mapping $h_{spda}$ of the SPDAgg layer is defined as:  

\begin{align}\label{eq:sumbimap}
\begin{split}
\mathbf{Y} & = h_{spda} ((\mathbf{X}_1,\ldots,\mathbf{X}_{N});\mathbf{W}_{1},\ldots,\mathbf{W}_{N}) \\ & = \sum_{i=1}^{N} \mathbf{W}_{i} \mathbf{X}_{i} (\mathbf{W}_{i})^T,
\end{split}
\end{align}
where $\mathbf{X}_{i} \in \mathbb{R}^{d^s_{in} \times d^s_{in}},i=1,\ldots,N$, are the input SPD matrices, 
$\mathbf{W}_{i} \in \mathbb{R}^{d^s_{out} \times d^s_{in}}$ are the transformation matrices,
$\mathbf{Y} \in \mathbb{R}^{d^s_{out} \times d^s_{out}}$ is the output matrix.

To guarantee that the output $\mathbf{Y}$ is SPD,
we remark that the right-hand side of Eq.~(\ref{eq:sumbimap}) can be rewritten as:
\begin{equation}
\sum_{i=1}^{N} \mathbf{W}_{i} \mathbf{X}_{i} (\mathbf{W}_{i})^T = \hat{\mathbf{W}} \diag(\mathbf{X}_{1},\ldots,\mathbf{X}_{N}) (\hat{\mathbf{W}})^T,
\end{equation}
where $\hat{\mathbf{W}} = [\mathbf{W}_{1},\ldots,\mathbf{W}_{N}]$ and $\diag(\mathbf{X}_{1},\ldots,\mathbf{X}_{N})$ 
is constructed such that its diagonal contains the diagonal entries of $\mathbf{X}_{1},\ldots,\mathbf{X}_{N}$:
\begin{equation}\label{eq:diag_cov}
\diag(\mathbf{X}_{1},\ldots,\mathbf{X}_{N}) = \begin{bmatrix} \mathbf{X}_{1} \ldots \ldots \ldots \\ \ldots \mathbf{X}_{2} \ldots \ldots \\ \ldots \ldots \ldots \mathbf{X}_{N}  \end{bmatrix}.
\end{equation}

It can be easily seen that $\diag(\mathbf{X}_{1},\ldots,\mathbf{X}_{N})$ is a valid SPD matrix, 
as for any vector $\mathbf{x} \neq 0$, one has $\mathbf{x}^T \diag(\mathbf{X}_{1},\ldots,\mathbf{X}_{N}) \mathbf{x} = \sum_{i=1}^{N} (\mathbf{x}_{i})^T \mathbf{X}_{i} \mathbf{x}_{i}$, where $\mathbf{x}=[(\mathbf{x}_{1})^T, \ldots, (\mathbf{x}_{N})^T]^T$ and the vectors $\mathbf{x}_{i},i=1,\ldots,N$, have equal sizes. The right-hand side of the above equation is strictly positive since $(\mathbf{x}_{i})^T \mathbf{X}_{i} \mathbf{x}_{i} \ge 0, 
\forall i=1,\ldots,N$, and there must exist
$i' \in \{1,\ldots,N\}$ such that $\mathbf{x}_{i'} \neq 0$ (as $\mathbf{x} \neq 0$), which implies that $(\mathbf{x}_{i'})^T \mathbf{X}_{i'} \mathbf{x}_{i'} > 0$.

Inspired by~\cite{HuangGool17}, we assume that the combined matrix 
$\hat{\mathbf{W}} = [\mathbf{W}_{1},\ldots,\mathbf{W}_{N}]$ is a full row rank matrix. 
Then optimal solutions of the transformation matrices are achieved by additionally assuming that
$\hat{\mathbf{W}}$ resides on a compact Stiefel manifold $St(d^s_{out},N \times d^s_{in})$\footnote{A compact Stiefel manifold $St(d^s_{out},N \times d^s_{in})$ is the set of $d^s_{out}$-dimensional orthonormal matrices of $\mathbb{R}^{N \times d^s_{in}}$.}. 
The transformation matrices $\mathbf{W}_{i}$ are updated by optimizing $\hat{\mathbf{W}}$
and projecting the optimal $\hat{\mathbf{W}}$ on its columns. 
Note that the constraint on the dimension $d^s_{out}$ of the output $\mathbf{Y}$ is: $d^s_{out} \le N d^s_{in}$.

\begin{figure}[t]
{
\centerline{\scalebox{0.116}{\input{spdc_net_t.tex}}}
\caption{\label{fig:spdc_network} The architecture of SPDC-NET. 
}}
\end{figure}

To map the output SPD matrix of the SPDAgg layer to an Euclidean space, we use the LogEig layer, followed
by a fully connected (FC) layer and a softmax layer.

\subsection{Gesture Recognition}
\label{subsec:gesture_recognition}

The SPDAgg layer outputs a matrix $\mathbf{B}\,{\in }\,\mathbb{R}^{d_{out}^s}$ for
each gesture sequence (see Fig.~\ref{fig:spdc_network}). This matrix is then transformed to its matrix logarithm and finally vectorized. The final representation of the gesture sequence is $\mathbf{v} = [b_{1,1},$ $\sqrt{2}b_{1,2}, \sqrt{2}b_{1,3},\ldots,\sqrt{2}b_{1,d_{out}^s}, b_{2,2}, \sqrt{2}b_{2,3}, \ldots, b_{d_{out}^s,d_{out}^s}]^T$ where $b_{i,i},i=1,\ldots,d_{out}^s$, are the diagonal entries of $\text{log}(\mathbf{B})$ and $b_{i,j},i<j,i,j=1,\ldots,d_{out}^s$, are the off-diagonal entries 
 of $\text{log}(\mathbf{B})$.

\subsection{Relation with Previous Works}
\label{subsec:links_previous_works}

Our approach is closely related to~\cite{HuangGool17,G2DeNet17}. 
We point out in the following paragraphs the relations between the proposed network and those introduced in~\cite{HuangGool17,G2DeNet17}. 

\begin{itemize}
\item Our network considers  temporal modeling for hand gesture recognition, while temporal modeling is not considered in~\cite{G2DeNet17} 
as they focus on image classification tasks.
Moreover, in our work, a Gaussian is identified as a SPD matrix, while~\cite{G2DeNet17} identifies a Gaussian as the square root of a SPD matrix. 
\item Our network takes directly 3D coordinates of hand joints as input, while in~\cite{HuangGool17},  covariance matrices must be computed beforehand
as input of their network. 
\item Our network relies not only on the second-order information (covariance) as~\cite{HuangGool17} but also on the first-order information (mean).
The first-order information has been proven to be useful in capturing the extra distribution information of low-level features~\cite{Sanchez13}. 
Moreover, we consider the first- and second-order information for different subsets of hand joints, 
while~\cite{HuangGool17} uses the whole set of joints to compute statistics.  
Our network is thus based on a finer granularity than~\cite{HuangGool17}.  
\item Our network combines two different and complementary architectures to better capture relevant statistics for the recognition task,
which makes our network distinct from those of~\cite{HuangGool17,G2DeNet17}. 
\end{itemize}

\section{Experiments}
\label{sec:exps} 

We conducted experiments using the Dynamic Hand Gesture (DHG) dataset~\cite{SmedtCVPRW16,SmedtEuro17} 
and the First-Person Hand Action (FPHA) dataset~\cite{Garcia-HernandoCVPR18}. In all experiments, the dimension of a output feature vector of the convolutional layer was set to $9$ ($d^c_{out}=9$), 
the dimensions of the transformation matrices of the SPDAgg layer were set to $200 \times 56$ ($d^s_{in}=56$, $d^s_{out}=200$). All sequences of the two datasets were normalized to have $500$ frames ($N_F=500$)\footnote{We tested with $N_F=300,500,800$ and the difference between obtained results were marginal.}.
The batch size and the learning rate were set to 30 and 0.01, respectively. 
The rectification threshold $\epsilon$ for the ReEig layer was set to 0.0001~\cite{HuangGool17}. 
The network trained at epoch $15$ was used to create the final gesture representation. The classifier was learned using the LIBLINEAR library~\cite{Fan08} with L2-regularized L2-loss (dual) where C was set to 1, 
the tolerance of termination criterion was set to 0.1 and no bias term was added.
For FPHA dataset, the non-optimized CPU implementation of our network on a 3.4GHz machine with 24GB
RAM and Matlab R2015b takes about 22 minutes per epoch and 7 minutes per epoch for training and testing, respectively. In the following, we provide details on the experimental settings and results obtained for each dataset.

\subsection{Datasets and Experimental Settings}
\label{subsec:dataset_experimental_setting}

{\bf DHG dataset.} 
The DHG dataset contains 14 gestures performed in two ways: using one finger and the whole hand. 
Each gesture is executed several times by different actors. 
Gestures are subdivided into fine and coarse categories. The dataset provides the 3D coordinates of 22 hand joints as illustrated in Fig.~\ref{fig:hand_graph}(a). It has been split into $1960$ train sequences (70\% of the dataset) and $840$ test sequences (30\% of the dataset)~\cite{SmedtEuro17}. 

{\bf FPHA dataset.} 
This dataset contains $1175$ action videos belonging to $45$ different action categories, in 3 different scenarios, and performed by 6 actors.
Action sequences present high inter-subject and intra-subject variability of style, speed, scale, and viewpoint. 
The dataset provides the 3D coordinates of 21 hand joints as DHG dataset except for the palm joint. We used the 1:1 setting proposed in~\cite{Garcia-HernandoCVPR18} with $600$ action sequences for training and $575$ for testing.

\begin{table}[t]
\begin{center}
  \resizebox{1.0\linewidth}{!}{
  \def\arraystretch{1.2}
  \begin{tabular}{l c  c  c }    
    \hline
    {\bf Num. of a hand joint's neighbors} & {\bf FPHA} & {\bf DHG (14 gestures)} & {\bf DHG (28 gestures)}  \\      
    \hline
    $3$  &  91.65  &  93.10  & 88.33 \\
    $9$  &  93.22  &  94.29  &  89.40  \\ 
    \hline
  \end{tabular}
  } 
\end{center}
\caption{\label{tab:exp_ablation_hand_modeling} Recognition accuracy (\%) of our network for different settings of hand joint's neighborhood.}
\end{table}

\begin{table}[t]
\begin{center}
  \resizebox{0.8\linewidth}{!}{
  \def\arraystretch{1.2}
  \begin{tabular}{l c  c  c }    
    \hline
    $t_0$ & {\bf FPHA} & {\bf DHG (14 gestures)} & {\bf DHG (28 gestures)}  \\      
    \hline
    $1$  &  93.22  & 94.29  &   89.40   \\
    $2$  &  93.04  & 94.17  &   89.04   \\ 
    $3$  &  93.04  & 94.29  &   89.40   \\ 
    \hline
  \end{tabular}
  } 
\end{center}
\caption{\label{tab:exp_ablation_t0} Recognition accuracy (\%) of our network for different settings of $t_0$.}
\end{table}

\subsection{Ablation Study}
\label{subsec:ablation_study}

In this section, we examine the influence of different components of our network on its accuracy. The default values of $t_0$ and $N_S$ are set to 1 and 15, respectively.

{\bf Hand modeling.}
We evaluate the performance of our network when only physical connections of hand joints are used for the computations at the convolutional layer, i.e., connections between hand joints belonging to neighboring fingers are removed from the graph in Fig.~\ref{fig:hand_graph} (b). Each joint is now connected to at most three joints including itself. Results shown in Tab.~\ref{tab:exp_ablation_hand_modeling} confirm that the use of connections other than physical connections of hand joints bring performance improvement. 

{\bf Time interval $t_0$.} 
In this experiment, we vary $t_0$ and keep other components of our network unchanged. To ensure that the computation of covariance matrices is numerically stable, we set $t_0>0$. Tab.~\ref{tab:exp_ablation_t0} shows the performance of our network with three different settings of $t_0$, i.e. $t_0=1,2,3$. Results suggest that using 3 consecutive frames for the input of the first GaussAgg layer of ST-GA-NET is sufficient to obtain good performance.

{\bf Number $N_S$ of sub-sequences in a branch.}
This experiment is performed by varying $N_S$ while keeping other components of our network unchanged. 
For the same reason related to the computation of covariance matrices, $N_S$ must be in a certain interval. We tested with $N_S=15,20,25$. Results given in Tab.~\ref{tab:exp_ablation_ns} indicate that our network is not sensitive to different settings of $N_S$.

{\bf Contribution of ST-GA-NET and TS-GA-NET.} 
We evaluate the performance of two networks, referred to as ST-HGR-NET and TS-HGR-NET by removing sub-networks TS-GA-NET and ST-GA-NET from our network, respectively. Results shown in Tab. \ref{tab:exp_ablation_contribs} reveal that none of both ST-GA-NET and TS-GA-NET always provides the best performances on the datasets. This motivates the need for their combination using the component SPDC-NET and this contributes to the overall performance of our global network combining both TS-GA-NET and ST-GA-NET. 

In the following, we report results obtained with default settings of $t_0$ and $N_S$, i.e. $t_0=1$ and $N_S=15$. 

\begin{table}[t]
\begin{center}
  \resizebox{0.8\linewidth}{!}{
  \def\arraystretch{1.2}
  \begin{tabular}{l c  c  c }    
    \hline
    $N_S$ & {\bf FPHA} & {\bf DHG (14 gestures)} & {\bf DHG (28 gestures)}  \\      
    \hline
    $15$  &  93.33  & 94.29  & 89.40  \\
    $20$  &  92.87  & 94.05  & 88.93     \\ 
    $25$  &  92.70  & 94.29  & 89.04     \\ 
    \hline
  \end{tabular}
  } 
\end{center}
\caption{\label{tab:exp_ablation_ns} Recognition accuracy (\%) of our network for different settings of $N_S$.}
\end{table}

\begin{table}[t]
\begin{center}
  \resizebox{0.9\linewidth}{!}{
  \def\arraystretch{1.2}
  \begin{tabular}{ l  c  c  c}    
    \hline
    {\bf Network} & {\bf FPHA} & {\bf DHG (14 gestures)} & {\bf DHG (28 gestures)} \\ 
    \hline        
    {\bf ST-HGR-NET} & 91.83  & 93.21  & 89.29 \\
    {\bf TS-HGR-NET} & 90.96 & 93.33 &  88.21  \\                  
    {\bf ST-TS-HGR-NET} & {\bf 93.22} & {\bf 94.29} & {\bf 89.40} \\    
    \hline	
  \end{tabular}
  } 
\end{center}
\caption{\label{tab:exp_ablation_contribs} Recognition accuracy (\%) of sub-networks ST-GA-NET and TS-GA-NET.}
\end{table}

\subsection{Comparison with State-of-the-Art}
\label{subsec:comparison_state_of_the_art}

{\bf DHG dataset.}
The comparison of our method and state-of-the-art methods on DHG dataset is given in Tab.~\ref{tab:exp_dhg}. 
The accuracy of the method of~\cite{HuangGool17} is obtained by using the implementation provided by the authors with their default parameter settings. 
Our method significantly outperforms the competing ones. The network of~\cite{HuangGool17} also learns a SPD matrix-based representation from skeletal data which is similar in spirit to our network. However, they concatenate the 3D coordinates of joints at each frame to create the feature vector of that frame, and their network's input is the covariance matrix computed from feature vectors over the whole skeleton sequence. Thus, spatial and temporal relationships of joints are not effectively taken into account. By exploiting these relationships, our network improves the recognition accuracy by 19.05\% and 19.76\% compared to the results of~\cite{HuangGool17} for experiments with 14 and 28 gestures, respectively. For more comparison of our method and existing methods, we conducted experiments using the leave-one-subject-out experimental protocol. Results on Tabs.~\ref{tab:exp_dhg_loo14} (14 gestures) and~\ref{tab:exp_dhg_loo28} (28 gestures) demonstrate that our method achieves the best results compared to existing methods on this protocol. 
In particular, our method outperforms the most recent work~\cite{WengTraversalConvECCV18} by 1.5 and 3 percent points for experiments with 14 and 28 gestures, respectively.

\begin{table}[t]
\begin{center}
  \resizebox{1.0\linewidth}{!}{
  \def\arraystretch{1.2}
  \begin{tabular}{ l  c  c  c  c  c  c}    
    \hline
    \multirow{2}{*}{{\bf Method}} & \multirow{2}{*}{{\bf Year}} & \multirow{2}{*}{{\bf Color}} & \multirow{2}{*}{{\bf Depth}} & \multirow{2}{*}{{\bf Pose}} & \multicolumn{2}{c}{{\bf Accuracy (\%)}} \\ 
    \cline{6-7}
    & & & & & {\bf 14 gestures} & {\bf 28 gestures}  \\          
    \hline    
    Oreifej and Liu~\cite{OreifejHon4d13} & 2013 & \xmark & \checkmark & \xmark & 78.53 & 74.03 \\
    Devanne et al.~\cite{DevanneRie15} & 2015 & \xmark & \xmark & \checkmark & 79.61 & 62.00 \\
    Huang et al.~\cite{HuangGool17} & 2017 & \xmark & \xmark & \checkmark & 75.24 & 69.64 \\
    Ohn-Bar and Trivedi~\cite{OhnBar2013} & 2013 & \xmark & \xmark & \checkmark & 83.85 & 76.53 \\
    Chen et al.~\cite{ChenDBLP17} & 2017 & \xmark & \xmark & \checkmark & 84.68 & 80.32 \\
    De Smedt et al.~\cite{SmedtCVPRW16} & 2016 & \xmark & \xmark & \checkmark & 88.24 & 81.90 \\
    Devineau et al.~\cite{DevineauFG18} & 2018 & \xmark & \xmark & \checkmark & 91.28 & 84.35 \\
    \hline
    {\bf ST-TS-HGR-NET} & & \xmark & \xmark & \checkmark & {\bf 94.29}  & {\bf 89.40}  \\         
    \hline	
  \end{tabular}
  } 
\end{center}
\caption{\label{tab:exp_dhg} Recognition accuracy comparison of our method and state-of-the-art methods on DHG dataset with 1960 sequences for training and 840 sequences for testing. The best result in each column is marked in bold.}
\end{table}


\begin{table}[t]
\begin{center}
  \resizebox{1.0\linewidth}{!}{
  \def\arraystretch{1.2}
  \begin{tabular}{l c  c  c  c  c}
    \hline    
     {\bf Method}  &  {\bf Year}  &  {\bf Color}  &  {\bf Depth}  &  {\bf Pose}  &  {\bf Accuracy (\%)}  \\ 
    \hline    
    De Smedt et al.,~\cite{SmedtCVPRW16} & 2016 & \xmark & \xmark & \checkmark &  83.1 \\         
    CNN+LSTM~\cite{Nez2018} & 2018 & \xmark & \xmark & \checkmark & 85.6 \\ 
    Weng et al.,~\cite{WengTraversalConvECCV18} & 2018 & \xmark & \xmark & \checkmark & 85.8 \\
    {\bf ST-TS-HGR-NET} & & \xmark & \xmark & \checkmark & {\bf 87.3} \\
    \hline
  \end{tabular}
  } 
\end{center}
\caption{\label{tab:exp_dhg_loo14} Recognition accuracy comparison of our method and state-of-the-art methods on DHG dataset 
using the leave-one-subject-out experimental protocol with $14$ gestures. The best result in each column is marked in bold.}
\end{table}

\begin{table}[t]
\begin{center}
  \resizebox{1.0\linewidth}{!}{
  \def\arraystretch{1.2}
  \begin{tabular}{l  c  c  c  c  c}
    \hline    
     {\bf Method}  &  {\bf Year}  &  {\bf Color}  &  {\bf Depth}  &  {\bf Pose}  &  {\bf Accuracy (\%)}  \\   
    \hline    
    De Smedt et al.,~\cite{SmedtCVPRW16} & 2016 & \xmark & \xmark & \checkmark & 80.0 \\         
    CNN+LSTM~\cite{Nez2018} & 2018 & \xmark & \xmark & \checkmark & 81.1 \\ 
    Weng et al.,~\cite{WengTraversalConvECCV18} & 2018 & \xmark & \xmark & \checkmark & 80.4 \\
    {\bf ST-TS-HGR-NET} & & \xmark & \xmark & \checkmark & {\bf 83.4} \\
    \hline
  \end{tabular}
  } 
\end{center}
\caption{\label{tab:exp_dhg_loo28} Recognition accuracy comparison of our method and state-of-the-art methods on DHG dataset 
using the leave-one-subject-out experimental protocol with $28$ gestures. The best result in each column is marked in bold.}
\end{table}

{\bf FPHA dataset.} 
Tab.~\ref{tab:exp_first_person_hand_action} shows the accuracies of our method and state-of-the-art methods on FPHA dataset. The accuracies of the methods of~\cite{HuangGool17} and~\cite{HuangAAAI18} are obtained by using the implementations provided by the authors with their default parameter settings. 
Despite the simplicity of our network compared to the competing deep neural networks, it is superior to them on this dataset. 
The best performing method among state-of-the-art methods is Gram Matrix, which gives 85.39\% accuracy, 7.83 percent points inferior to our method. 
The remaining methods are outperformed by our method by more than 10 percent points. We observe that the method of~\cite{HuangGool17} performs well on this dataset. However, since this method does not fully exploit spatial and temporal relationships of skeleton joints, it gives a significantly lower accuracy than our method.
Results again confirm the effectiveness of the proposed network architecture for hand gesture recognition.

\begin{table}[t]
\begin{center}
  \resizebox{0.95\linewidth}{!}{
  \def\arraystretch{1.2}
  \begin{tabular}{ l  c  c c c c}    
    \hline
    {\bf Method} & {\bf Year} & {\bf Color} & {\bf Depth} & {\bf Pose} & {\bf Accuracy (\%)}  \\          
    \hline    
    Two stream-color~\cite{Feichtenhofer2016ConvolutionalTN} & 2016 & \checkmark & \xmark & \xmark & 61.56 \\             
    Two stream-flow~\cite{Feichtenhofer2016ConvolutionalTN} & 2016 & \checkmark & \xmark & \xmark & 69.91 \\         
    \hline	
    Two stream-all~\cite{Feichtenhofer2016ConvolutionalTN} & 2016 & \checkmark & \xmark & \xmark & 75.30 \\         
    \hline	
    HOG$^2$-depth~\cite{Ohn-BarHandAuto14} & 2013 & \xmark & \checkmark & \xmark & 59.83 \\         
    HOG$^2$-depth+pose~\cite{Ohn-BarHandAuto14} & 2013 & \xmark & \checkmark & \checkmark & 66.78 \\
    HON4D~\cite{OreifejHon4d13} & 2013 & \xmark & \checkmark & \xmark & 70.61 \\                        
    Novel View~\cite{Rahmani16} & 2016 & \xmark & \checkmark & \xmark & 69.21 \\
    \hline	
    1-layer LSTM~\cite{ZhuCo-occu16} & 2016 & \xmark & \xmark & \checkmark & 78.73 \\
    2-layer LSTM~\cite{ZhuCo-occu16} & 2016 & \xmark & \xmark & \checkmark & 80.14 \\                        
    \hline
    Moving Pose~\cite{ZanfirMovingPose13} & 2013 & \xmark & \xmark & \checkmark & 56.34 \\
    Lie Group~\cite{LieGroup14} & 2014 & \xmark & \xmark & \checkmark & 82.69 \\
    HBRNN~\cite{NN15} & 2015 & \xmark & \xmark & \checkmark & 77.40 \\
    Gram Matrix~\cite{ZhanGramMatrix16} & 2016 & \xmark & \xmark & \checkmark & 85.39 \\
    TF~\cite{GarciaHernando2016TransitionFL} & 2017 & \xmark & \xmark & \checkmark & 80.69 \\
    \hline
    JOULE-color~\cite{HuJointHeter15} & 2015 & \checkmark & \xmark & \xmark & 66.78 \\
    JOULE-depth~\cite{HuJointHeter15} & 2015 & \xmark & \checkmark & \xmark & 60.17 \\
    JOULE-pose~\cite{HuJointHeter15} & 2015 & \xmark & \xmark & \checkmark & 74.60 \\
    \hline
    JOULE-all~\cite{HuJointHeter15} & 2015 & \checkmark & \checkmark & \checkmark & 78.78 \\
    \hline
    Huang et al.~\cite{HuangGool17} & 2017 & \xmark & \xmark & \checkmark & 84.35 \\
    \hline
    Huang et al.~\cite{HuangAAAI18} & 2018 & \xmark & \xmark & \checkmark & 77.57 \\
    \hline
    {\bf ST-TS-HGR-NET} & & \xmark & \xmark & \checkmark & {\bf 93.22}  \\
    \hline
  \end{tabular}
  } 
\end{center}
\caption{\label{tab:exp_first_person_hand_action} Recognition accuracy comparison of our method and state-of-the-art methods on FPHA dataset. 
The best result in each column is marked in bold.}
\end{table}

\section{Conclusion}
\label{sec:conclu}
We have presented a new neural network for hand gesture recognition that learns a discriminative SPD matrix encoding the first-order and second-order statistics. 
We have provided the experimental evaluation on two benchmark datasets showing that our method outperforms state-of-the-art methods.


\paragraph*{Acknowledgments.} This material is based upon work supported by the European Union and the Region Normandie under the project IGIL. We thank Guillermo Garcia-Hernando for providing access to FPHA dataset~\cite{Garcia-HernandoCVPR18}.

{\small
\bibliographystyle{ieee}
\bibliography{references}
}


\begin{appendices}
\section{Backpropagation Procedures}\label{appendix}
This part provides details on the backpropagation procedures during the training process of our network. Our network can be encoded as a pair $(h,\mathbf{W})$ where $h = h^{(N_L)} \circ \ldots \circ h^{(1)}$
is a composition of $N_L$ layers, $\mathbf{W} = (\mathbf{W}({N_L}),\ldots,\mathbf{W}(1))$ represents the network parameters, $\mathbf{W}(k)$ are the parameters of layer $k$.
Let $L^{(k)} = L \circ h^{(N_L)} \circ \ldots \circ h^{(k)}$ be the loss as a function of layer $k-1$.
In the following, we omit the superscript $k$ of $L^{(k)}$ for the sake of convenience. 

\subsection{SPDAgg layer}
\label{subsec:backprop_concatbimap}

We present in this section a method based on the chain rule of~\cite{Ionescu2015} for the computations of partial derivatives. For more details on the established theory, we refer readers to~\cite{Ionescu2015}. The variation of $Y$ is given by:
\begin{align}\label{eq:dx}
\begin{split}
d\mathbf{Y} = & \sum_{i=1}^{N} \Big( d\mathbf{W}_i \mathbf{X}_i (\mathbf{W}_i)^T + \mathbf{W}_i d\mathbf{X}_i (\mathbf{W}_i)^T \\ & + \mathbf{W}_i \mathbf{X}_i d(\mathbf{W}_i)^T \Big).
\end{split}
\end{align}

The chain rule in this case is:
\begin{equation}\label{eq:chain_rule}
\frac{\partial L}{\partial \mathbf{Y}} : d\mathbf{Y} = \sum_{i=1}^{N} \Big( \frac{\partial L}{\partial \mathbf{W}_i} : d\mathbf{W}_i + \frac{\partial L}{\partial \mathbf{X}_i} : d\mathbf{X}_i \Big).
\end{equation}

By replacing $d\mathbf{Y}$ in Eq.~(\ref{eq:chain_rule}) with its expression in Eq.~(\ref{eq:dx}), the left-hand side of Eq.~(\ref{eq:chain_rule}) becomes:
\begin{align}\label{eq:substitue}
\begin{split}
& \sum_{i=1}^{N} \Big( \frac{\partial L}{\partial \mathbf{Y}} : d\mathbf{W}_i \mathbf{X}_i (\mathbf{W}_i)^T + \frac{\partial L}{\partial \mathbf{Y}} : \mathbf{W}_i \mathbf{X}_i d(\mathbf{W}_i)^T \\ & + \frac{\partial L}{\partial \mathbf{Y}} : \mathbf{W}_i d\mathbf{X}_i (\mathbf{W}_i)^T \Big).
\end{split}
\end{align}

Using the properties~\cite{Ionescu2015} of the matrix inner product ``:'' and by the fact that $\mathbf{Y}$ and $\mathbf{X}_{i}$ are symmetric, we have:
\begin{equation}
\frac{\partial L}{\partial \mathbf{Y}} : d\mathbf{W}_i \mathbf{X}_i (\mathbf{W}_i)^T = \frac{\partial L}{\partial \mathbf{Y}} \mathbf{W}_i \mathbf{X}_i : d\mathbf{W}_i,
\end{equation}
\begin{equation}
\frac{\partial L}{\partial \mathbf{Y}} : \mathbf{W}_i \mathbf{X}_i d(\mathbf{W}_i)^T = \frac{\partial L}{\partial \mathbf{Y}} \mathbf{W}_i \mathbf{X}_i : d\mathbf{W}_i,
\end{equation}
\begin{equation}
\frac{\partial L}{\partial \mathbf{Y}} : \mathbf{W}_i d\mathbf{X}_i (\mathbf{W}_i)^T = (\mathbf{W}_i)^T \frac{\partial L}{\partial \mathbf{Y}} \mathbf{W}_i : d\mathbf{X}_i.
\end{equation}

The expression~(\ref{eq:substitue}) now becomes:
\begin{equation}
\sum_{i=1}^{N} \Big( 2\frac{\partial L}{\partial \mathbf{Y}} \mathbf{W}_i \mathbf{X}_i : d\mathbf{W}_i + (\mathbf{W}_i)^T \frac{\partial L}{\partial \mathbf{Y}} \mathbf{W}_i : d\mathbf{X}_i \Big).
\end{equation}

Since the last expression is equal to the right-hand side of Eq.~(\ref{eq:chain_rule}), we obtain the partial derivatives:
\begin{equation}\label{eq:weight_gradient}
\frac{\partial L}{\partial \mathbf{W}_i} = 2\frac{\partial L}{\partial \mathbf{Y}} \mathbf{W}_i \mathbf{X}_i,
\end{equation}
\begin{equation}
\frac{\partial L}{\partial \mathbf{X}_i} = (\mathbf{W}_i)^T \frac{\partial L}{\partial \mathbf{Y}} \mathbf{W}_i.
\end{equation}

To learn the weights of this layer, we use the method proposed in~\cite{HuangGool17}. 
The weight $\hat{\mathbf{W}}$ is updated in two steps. First, the tangential component to the Stiefel manifold
is obtained by subtracting the normal component of the Euclidean gradient:

\begin{equation}
\tilde{\nabla} L_{\hat{\mathbf{W}}^t} = \nabla L_{\hat{\mathbf{W}}^t} - \nabla L_{\hat{\mathbf{W}}^t} (\hat{\mathbf{W}}^t)^T \hat{\mathbf{W}}^t,
\end{equation}
where $\hat{\mathbf{W}}^t$ is the updated weight at the $t^{th}$ iteration and 
$\nabla L_{\hat{\mathbf{W}}^t} (\hat{\mathbf{W}}^t)^T \hat{\mathbf{W}}^t$ is the normal component
of the Euclidean gradient $\nabla L_{\hat{\mathbf{W}}^t}$. 
Following Eq.~(\ref{eq:weight_gradient}), the Euclidean gradient $\nabla L_{\hat{\mathbf{W}}^t}$ is given by:
\begin{equation}
\nabla L_{\hat{\mathbf{W}}^t} =  2\frac{\partial L}{\partial \mathbf{Y}} [\text{Proj}_1(\hat{\mathbf{W}}^t) \mathbf{X}_1,\ldots,\text{Proj}_{N}(\hat{\mathbf{W}}^t) \mathbf{X}_N],
\end{equation}
where $\text{Proj}_i(\hat{\mathbf{W}}^t),i=1,\ldots,N$, is the projection of $\hat{\mathbf{W}}^t$ on its
columns corresponding to $\mathbf{W}_i$.

Then a retraction operation is used to map back the updated weight in the tangent space of the Stiefel manifold
to that in the Stiefel manifold as:

\begin{equation}\label{eq:bp_weight_update_2}
\hat{\mathbf{W}}^{t+1} = \Gamma (\hat{\mathbf{W}}^t - \lambda \tilde{\nabla} L_{\hat{\mathbf{W}}^t}),
\end{equation}
where $\Gamma$ is the retraction operation, $\lambda$ is the learning rate.

The updated weights of $\mathbf{W}_i,i=1,\ldots,N$, at the $(t+1)^{th}$ iteration can be computed as:

\begin{equation}\label{eq:bp_submatrix_update}
(\mathbf{W}_i)^{t+1} = \text{Proj}_i(\hat{\mathbf{W}}^{t+1}).
\end{equation}

\subsection{LogEig and ReEig layers}
\label{subsec:backprop_other}

To make this document self-contained for readers, we present here the computations of partial derivatives
for the LogEig and ReEig layers. For more details, we refer readers to~\cite{HuangGool17,Ionescu2015}. 
For the LogEig layers, the first step receives matrix $\mathbf{X}^t_{s,f}$ as input and produces matrices $\mathbf{U}$ and $\mathbf{V}$ such that $\mathbf{X}^t_{s,f} = \mathbf{U} \mathbf{V} \mathbf{U}^T$.  
The partial derivatives $\frac{\partial L}{\partial \mathbf{X}^t_{s,f}}$ can be computed from 
those of the outputs $\frac{\partial L}{\partial \mathbf{U}}$ and $\frac{\partial L}{\partial \mathbf{V}}$ as~\cite{Ionescu2015}:
\begin{equation}\label{eq:eigen_decom}
\frac{\partial L}{\partial \mathbf{X}^t_{s,f}} = \mathbf{U} \Bigg \{ 2 \bigg( \tilde{K}^T \circ \Big( \mathbf{U}^T \frac{\partial L}{\partial \mathbf{U}} \Big)_{sym} \bigg) + \Big( \frac{\partial L}{\partial \mathbf{V}} \Big)_{diag} \Bigg \} \mathbf{U}^T,
\end{equation}
where $\mathbf{D}_{sym} = \frac{1}{2} (\mathbf{D} + \mathbf{D}^T)$, $\mathbf{D}_{diag}$ is $\mathbf{D}$ with all off-diagonal elements being 0, 
and $\tilde{K}^T$ is defined as:
\begin{equation}
\tilde{K}_{ij} = \begin{cases} \frac{1}{\sigma_i - \sigma_j}, & i \neq j \\ 0, & i=j \end{cases}
\end{equation}

The second step receives matrices $\mathbf{U}$ and $\mathbf{V}$ as input 
and produces matrix $\mathbf{Y}^t_{s,f} = \mathbf{U} \log(\mathbf{V}) \mathbf{U}^T$. 
The partial derivatives $\frac{\partial L}{\partial \mathbf{U}}$ and $\frac{\partial L}{\partial \mathbf{V}}$ 
can be computed from those of the output $\frac{\partial L}{\partial \mathbf{Y}^t_{s,f}}$ as~\cite{HuangGool17}:
\begin{equation}
\frac{\partial L}{\partial \mathbf{U}} = 2 \Big( \frac{\partial L}{\partial \mathbf{Y}^t_{s,f}} \Big)_{sym} \mathbf{U} \log(\mathbf{V}),
\end{equation}

\begin{equation}
\frac{\partial L}{\partial \mathbf{V}} = \mathbf{V}^{-1} \mathbf{U}^T \Big( \frac{\partial L}{\partial \mathbf{Y}^t_{s,f}} \Big)_{sym} \mathbf{U}.
\end{equation}

The ReEig layers can be decomposed into two steps as the LogEig layers where the partial derivatives of the first step are computed similarly to the LogEig layers. 
For the second step, the partial derivatives $\frac{\partial L}{\partial \mathbf{U}}$ and $\frac{\partial L}{\partial \mathbf{V}}$ 
can be computed from those of the output $\frac{\partial L}{\partial \mathbf{Y}^t_{s,f}}$ as:

\begin{equation}
\frac{\partial L}{\partial \mathbf{U}} = 2 \Big( \frac{\partial L}{\partial \mathbf{Y}^t_{s,f}} \Big)_{sym} \mathbf{U} \max(\epsilon \mathbf{I},\mathbf{V}),
\end{equation}

\begin{equation}
\frac{\partial L}{\partial \mathbf{V}} = \mathbf{Q} \mathbf{U}^T \Big( \frac{\partial L}{\partial \mathbf{Y}^t_{s,f}} \Big)_{sym} \mathbf{U},
\end{equation}
where $\max(\epsilon \mathbf{I},\mathbf{V})$ is defined in Eq.~(9), and $\mathbf{Q}$ is the gradient of 
$\max(\epsilon \mathbf{I},\mathbf{V})$ with diagonal elements being defined as:
\begin{equation}
\mathbf{Q}(i,i) = \left\lbrace\begin{array}{ll} 1 & \text{if } \mathbf{V}(i,i) > \epsilon \\ 0 & \text{if } \mathbf{V}(i,i) \le \epsilon. \end{array}\right.
\end{equation}

\subsection{VecMat layer}
\label{subsec:backprop_vecmat}

For the VecMat layer, the expression for the partial derivatives $\frac{\partial L}{\partial \mathbf{X}^t_{s,f}}$ 
is straightforward and can be written as:

\begin{align}
\begin{split}
& \frac{\partial L}{\partial \mathbf{X}^t_{s,f}} = \\ & \begin{bmatrix} \frac{\partial L}{\partial \mathbf{y}^t_{s,f}(1)} & \frac{\sqrt 2 \partial L}{\partial \mathbf{y}^t_{s,f}(2)} & \ldots & \frac{\sqrt 2 \partial L}{\partial \mathbf{y}^t_{s,f}(d^c_{out}+1)} \\ \frac{\sqrt 2 \partial L}{\partial \mathbf{y}^t_{s,f}(2)} & \frac{\partial L}{\partial \mathbf{y}^t_{s,f}(d^c_{out}+2)} & \ldots & \frac{\sqrt 2 \partial L}{\partial \mathbf{y}^t_{s,f}(2d^c_{out}+1)} \\ \frac{\sqrt 2 \partial L}{\partial \mathbf{y}^t_{s,f}(d^c_{out}+1)} & \frac{\sqrt 2 \partial L}{\partial \mathbf{y}^t_{s,f}(2d^c_{out}+1)} & \ldots & \frac{\partial L}{\partial \mathbf{y}^t_{s,f}(\frac{(d^c_{out}+1)(d^c_{out}+2)}{2})} \end{bmatrix},
\end{split}
\end{align}
where $\mathbf{y}^t_{s,f} = [\mathbf{y}^t_{s,f}(1),\ldots,\mathbf{y}^t_{s,f}(\frac{(d^c_{out}+1)(d^c_{out}+2)}{2})]^T$.

\subsection{GaussAgg layer}
\label{subsec:backprop_gaussagg}

The general form for the mapping $h_{ga}$ of the GaussAgg layers can be written as:

\begin{equation}\label{eq:gaussagg_mapping}
h_{ga}(\mathbf{X}) = \mathbf{Y} = \begin{bmatrix}  \pmb{\Sigma} + \pmb{\mu}\pmb{\mu}^T & \pmb{\mu} \\ \pmb{\mu}^T & 1 \end{bmatrix},
\end{equation}
where $\mathbf{X} = [\mathbf{x}_1,\ldots,\mathbf{x}_N]^T \in \mathbb{R}^{N \times d}$ is the input of the GaussAgg layer, 
$\mathbf{Y}$ is the output of the GaussAgg layer, $\pmb{\mu} = \frac{1}{N} \sum_{i=1}^{N} \mathbf{x}_i$ and 
$\pmb{\Sigma} = \frac{1}{N} \sum_{i=1}^{N} (\mathbf{x}_i - \pmb{\mu})(\mathbf{x}_i - \pmb{\mu})^T$.

By the identity $\pmb{\Sigma} = \frac{1}{N} \mathbf{X}^T \mathbf{X} - \pmb{\mu} \pmb{\mu}^T$,
$\mathbf{Y}$ can be expressed as a function of $\mathbf{X}$ as~\cite{G2DeNet17}:

\begin{equation}\label{eq:matrix_partition_analytic}
\mathbf{Y} = \frac{1}{N} \mathbf{B} \mathbf{X}^T \mathbf{X} \mathbf{B}^T + \frac{2}{N} \Big(\mathbf{B} \mathbf{X}^T \mathbf{1} \mathbf{b}^T\Big)_{sym} + \mathbf{C},
\end{equation}
where $\mathbf{B} = \begin{bmatrix} \mathbf{I} \\ \mathbf{0}^T \end{bmatrix}$, $\mathbf{I}$ is the $d \times d$ identity matrix
and $\mathbf{0}$ is the $d$-dimensional zero vector, $\mathbf{b} = [0,\ldots,0,1]^T$ is the $(d+1)$-dimensional vector with all elements
being zero except the last one which is equal to one, $\mathbf{1}$ is the $N$-dimensional vector with all elements being one, 
$\mathbf{C} = \begin{bmatrix} \mathbf{O} & \mathbf{0} \\ \mathbf{0}^T & \mathbf{1} \end{bmatrix}$, $\mathbf{O}$ is the $d \times d$ zero matrix.

Based on Eq.~(\ref{eq:matrix_partition_analytic}), the expression for the partial derivatives $\frac{\partial L}{\partial \mathbf{X}}$ can be obtained as:
\begin{equation}\label{eq:bp_matpartition}
\frac{\partial L}{\partial \mathbf{X}} = \frac{2}{N}\Big(\mathbf{X}\mathbf{B}^T + \mathbf{1}\mathbf{b}^T\Big)\Big(\frac{\partial L}{\partial \mathbf{Y}}\Big)_{sym}\mathbf{B}.
\end{equation}

\end{appendices}

\end{document}